%% file: paper.tex
\documentclass[10pt,twocolumn,letterpaper]{article}

\usepackage{cvpr}
\usepackage{times}
\usepackage{epsfig}
\usepackage{graphicx}
\usepackage{amsmath}
\usepackage{amssymb}
\usepackage{caption}
\usepackage{blindtext}
\usepackage{algpseudocode}
\usepackage{algorithm}

\newcommand{\mysubsubsection}[1]{\vspace{0.1cm} \noindent \underline{{\bf #1}}:}

\newcommand{\erik}[1]{}
\newcommand{\yasu}[1]{}

\usepackage[pagebackref=true,breaklinks=true,letterpaper=true,colorlinks,bookmarks=false]{hyperref}

\cvprfinalcopy 

\def\cvprPaperID{111} 

\ifcvprfinal\fi
\begin{document}


\title{Exploiting 2D Floorplan for Building-scale Panorama RGBD Alignment}

\author{Erik Wijmans \hspace*{0.5in} Yasutaka Furukawa\\
Washington University in St. Louis\\
{\tt\small \{erikwijmans, furukawa\}@wustl.edu}
}

\twocolumn[{%
\renewcommand\twocolumn[1][]{#1}%
\maketitle
\vspace{-1cm}
\begin{center}
 \includegraphics[width=\textwidth]{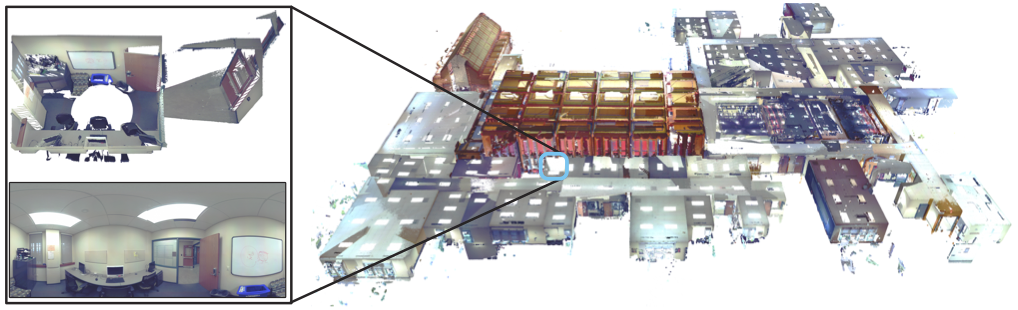}
 \captionof{figure}{The paper tackles building-scale panorama RGBD image
 alignment. Our approach utilizes a floorplan image to significantly
 reduce the number of necessary scans and hence human operating costs.
 } \label{fig:teaser}
\end{center}%
}]

\maketitle


\begin{abstract}
This paper presents a novel algorithm that utilizes a 2D floorplan to
align panorama RGBD scans.
While effective panorama RGBD alignment techniques exist, such a system
requires extremely dense RGBD image sampling. Our approach can
significantly reduce the number of necessary scans with the aid of a
floorplan image. We formulate a novel Markov Random Field inference
problem as a scan placement over the floorplan, as opposed to the
conventional scan-to-scan alignment. The technical contributions lie in
multi-modal image correspondence cues (between scans and schematic
floorplan) as well as a novel coverage potential avoiding an inherent
stacking bias.  The proposed approach has been evaluated on five
challenging large indoor spaces. To the best of our knowledge, we
present the first effective system that utilizes a 2D floorplan image
for building-scale 3D pointcloud alignment. The source code and the data
will be shared with the community to further enhance indoor mapping
research.
\end{abstract}

\input{introduction}
\input{related_work}

\input{input}

\input{model}

\input{algorithm}

\input{results}
\input{conclusion}
\input{acknowledgments}

\clearpage

{\small
\bibliographystyle{ieee}
\bibliography{bib}
}

\end{document}

%% file: introduction.tex
\section{Introduction}

\begin{figure*}[!h]
 \includegraphics[width=\textwidth]{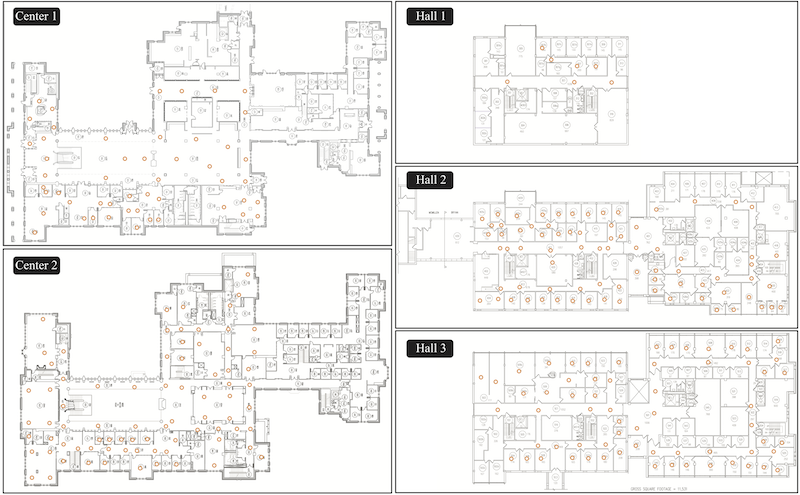}
\vspace{-0.62cm}
 \caption{We have used a professional grade laser range finder to
 acquire building-scale panorama RGBD scans over five floors in two
 buildings. An orange circle indicates a rough scan location.  Only one
 scan has been acquired in each room in most cases, making the
 intersection of adjacent scans minimal and the use of floorplan-image
 essential for our problem.
 }
 \label{fig:input}
\end{figure*}

3D scanning hardware has made remarkable progress in recent years, where
successful products exist in industry for commercial
applications. In particular, Panorama RGBD scanners have found
real-world applications as the system produces both 3D geometry
and immersive panorama images.
For instance, Faro 3D~\cite{faro3d} is a professional grade panorama
RGBD scanner, which can reach more than 100 meters and produce 100
million points per scan within a millimeter accuracy. The device is
perfect for 3D measurement, documentation, or surveillance in indoor
mapping, civil engineering or GIS applications.
Matterport~\cite{matterport,armeni20163d} is an emerging low-end
solution that can reach only 5 meters, but is much quicker (i.e., 1 to 2
minutes per scan), and has demonstrated compelling results for Real
Estate markets.

Given the success on the 3D scanning hardware, automated panorama RGBD
alignment has become a crucial technology. The Matterport system
provides a robust solution but requires extremely dense scanning (e.g.,
one scan every 2 to 3 meters). Dense scanning becomes infeasible for
high-end scanners (e.g., Faro 3D~\cite{faro3d}), whose single scan could
take thirty minutes or an hour depending on the resolution.  However,
these scanners are the only option for large buildings such as
department stores, airport terminals, or hotel lobbies, simply due to
the required operating ranges (e.g., 20 to 30 meters). Furthermore, the
precision of these high-end scanners is necessary for quantitative
recovery of metric information for scientific and engineering data
analysis.
In practice, with high-end 3D scanning devices, people use calibration
objects such as big bright balls and/or utilize semi-automatic 3D
alignment tool such as Autodesk ReCap 360~\cite{autodesk_recap} to
minimize the number of necessary scans.

%

This paper focuses on high-end 3D indoor scanning (See
Fig.~\ref{fig:teaser}). A key observation is that building targets for
high-end 3D scanning often have 2D floorplans. Our approach can
significantly reduce the number of necessary RGBD scans with the aid of
a 2D floorplan image.  The key technical contribution lies in a novel
Markov Random Field formulation as a scan placement problem as opposed
to the conventional scan-to-scan alignment. Besides the standard visual
and geometric feature matching between scans, we incorporate multi-modal
geometric or semantic correspondence cues associating scans and a
floorplan, as well as a novel ``coverage potential'' that avoids an
inherent {\it stacking bias}.
%
%
%
We have experimented with five challenging large indoor
spaces and demonstrated near perfect alignment results, significantly
outperforming existing approaches (See Fig.~\ref{fig:input}).
While our work has focused on existing indoor structures, the technology
can also be transformative to other domains, for example, Civil
Engineering applications at construction sites for progress monitoring
or safety inspection~\cite{karsch2014constructaide}, where precise
building blueprints exist.


%% file: related_work.tex
\section{Related work}

Two approaches exist for indoor 3D scanning: ``RGBD streaming'' or
``Panorama RGBD scanning''. RGBD streaming continuously moves a depth
camera and scans a scene. This has been the major choice among Computer
Vision
researchers~\cite{niessner2013real,choi2015robust,zhou2013elastic} after
the success of Kinect Fusion~\cite{izadi2011kinectfusion}. The input is
a RGBD video stream, where Simultaneous Localization and Mapping (SLAM)
is the core
technology.
Panorama RGBD scanning has been rather successful in industry, because
1) data acquisition is easy (i.e., picking a 2D position as opposed to 6
DoF navigation in RGBD streaming); 2) alignment is easier thanks to the
panoramic field of views; and 3) the system produces panorama images,
essential for many visualization applications. Structure from Motion
(SfM) is the core technology in this approach.
%
This paper provides an automated solution for Panorama RGBD alignment,
and the remainder of the section focuses on the description of the SfM
techniques, where we refer the reader to a survey
article~\cite{cadena2016simultaneous} for the SLAM literature.

Structure from Motion (SfM) addresses the problem of automatic image
alignment~\cite{hartley2003multiple}. State-of-the-art SfM system can
handle even millions of unorganized Internet
photographs~\cite{agarwal2011building,frahm2010building}. The wider field-of-view (e.g., panorama
images) further makes the alignment robust~\cite{shakernia2003structure,pagani2011structure,klingner2013street} as more
features are visible. When depth data is available, geometry provides
additional cues for alignment, where Iterative Closest Point
(ICP)~\cite{besl1992method} has been one of the most successful methods.
However, even state-of-the-art SfM or ICP systems face real challenges
for indoor scenes that are full of textureless walls with limited
visibility through narrow doorways. Existing approaches either 1) take
extremely dense samples~\cite{matterport} or 2) rely on manual
alignment~\cite{mura2014automatic,ikehata2015structured}.
%
%
%
%
%
%
%

The idea of utilizing building information for scan alignment
has been demonstrated in Civil Engineering
applications~\cite{karsch2014constructaide}.  However, the system
requires a full 3D model for construction design and
planning, as well as manual image-to-model correspondences to start the
process.
This paper utilizes 2D floorplan to align undersampled building-scale
panorama RGBD scans with minimal user interactions (i.e., a few mouse
strokes on a floorplan image).

%% file: input.tex
\section{Building-scale Panorama RGBD Datasets}

\begin{figure}[tb]
\centering
 \includegraphics[width=0.92\columnwidth]{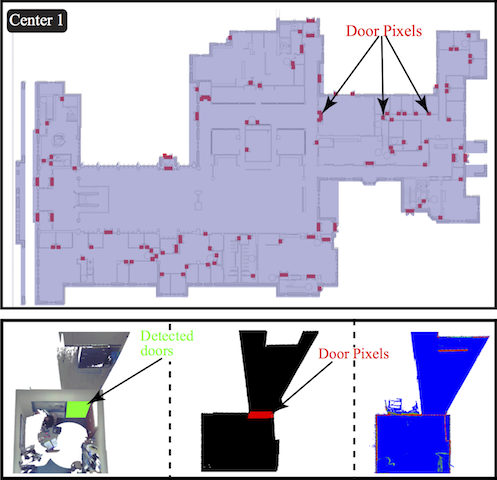}
 \vspace{-0.2cm}
 \caption{[Top] A floorplan image for Center 1 after the clutter
 removal. The blue overlay shows the building mask, and red pixels show
 the detected door pixels. [Bottom] The left shows the close-up of one
 scan with the result of 3D door detections. The middle shows the
 free-space image mask with detected door pixels. The right shows the
 point-evidence image with a heat-map color scheme
 }
 \label{fig:evidence}
\end{figure}

We have used a professional grade laser range finder, Faro
3D~\cite{faro3d}, to acquire panorama RGBD scans over five floors in two
buildings (Fig.~\ref{fig:input} and Table~\ref{table:stats}). A
floorplan is given as a rasterized image.  This section summarizes
standard preprocessing steps, necessary to prepare our datasets for our
algorithm. We here briefly describe these steps and refer the details to
the supplementary material (See Fig.~\ref{fig:evidence}).


\begin{table}[tb]
 \includegraphics[width=\columnwidth]{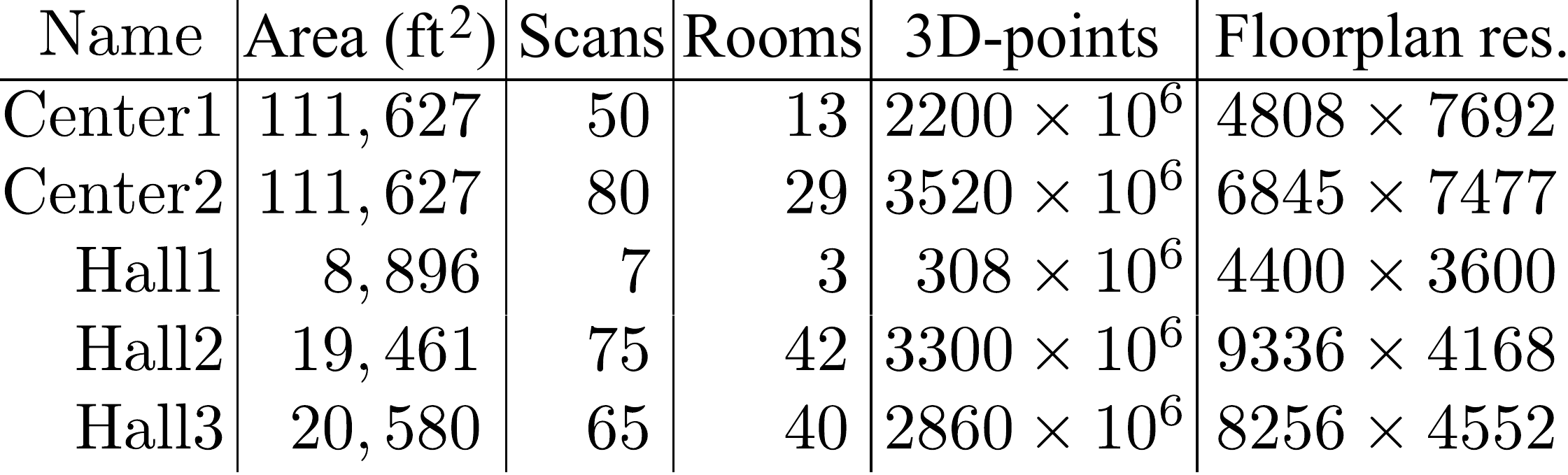}
 \vspace{-0.62cm}
 \caption{Statistics of our building-scale panorama RGBD datasets.
A single scan contains 44 million colored points.}
\label{table:stats}
\end{table}

\vspace{0.05cm}
\noindent $\bullet$ We remove clutter in a floorplan image by discarding small
connected components of black pixels.

\vspace{0.05cm}
\noindent $\bullet$
Our floorplan image contains a scale ruler, which lets us calculate
a metric scale per pixel with a few mouse strokes.
The process may be imprecise, and our algorithm will be designed to
tolerate errors.
In the absence of a ruler, we can align one scan with a
floorplan by hand to obtain a scale.

\vspace{0.05cm}
\noindent $\bullet$ We extract a
Manhattan frame from each scan and a floorplan image, respectively. In
each scan, we identify the vertical direction and the floor height
based on the point density.
%

\vspace{0.05cm}
\noindent $\bullet$ We turn each scan into {\it point} or {\it
free-space} evidence images in a top-down view. A point evidence
is a score in the range of $[0, 1]$, while a free-space evidence
is a binary mask.

\vspace{0.05cm}
\noindent $\bullet$ We compute a building mask over a floorplan image,
which can quickly prune unlikely scan placements. A
floorplan pixel becomes inside the mask if the pixel is between the left
and the right most pixels in its row and between the top and the bottom
most pixels in its column.

\vspace{0.05cm}
\noindent $\bullet$
We detect doors both in a floorplan image and 3D scans. For a floorplan,
we manually specify a bounding box containing a door symbol, and use a
standard template matching. For 3D scans, we use a heuristic to identify
door-specific 3D patterns directly in 3D points.

%
%




%% file: model.tex
\section{MRF formulation}

\begin{figure*}[tb]
  \centering
 \includegraphics[width=\textwidth]{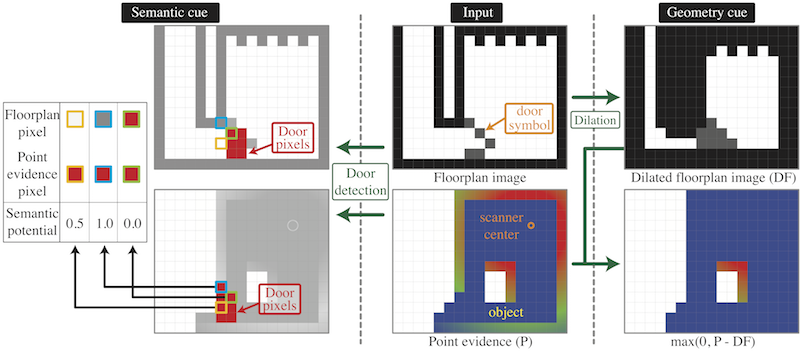}
 \vspace{-0.6cm}
 \caption{The scan-to-floor potential measures the consistency of the
 floorplan and a 3D scan in two ways. Left: The semantic cue
 checks the consistency of door detections over the door-detected pixels
 in the point evidence. Right: The geometric cue measures how much
 of the point evidence ($P$) is NOT explained by the dilated floorplan
 image.
%
 %
 }  \label{fig:scan-to-floor}
\end{figure*}

The multi-modal nature of the problem makes our formulation
fundamentally different from existing
ones~\cite{theiler2014fast,yan2016block}.
The first critical difference lies in the definition of the
variables. In existing approaches, a variable encodes a {\it 3D relative
placement} between a pair of scans~\cite{theiler2014fast,yan2016block}. In our
formulation, a variable encodes a {\it 2D absolute placement} of a
single scan over a floorplan image.



Let $\mathcal{S} = \{s_1, s_2, \cdots \}$ be our variables, where $s_i$
encodes the 2D placement of a single scan. $s_i$ consists of two
components: 1) rotation,
which takes one of the four angle values by exploiting the Manhattan frames
(0, 90, 180, or 270 degrees); and 2) translation,
which is a pixel coordinate in the floorplan image.
We seek to find the placements $\mathcal{S}$ that minimizes the
following three potentials:
\begin{eqnarray*}
 \sum_{s_i \in \mathcal{S}} E_{S}(s_i) +
  \sum_{(s_i,s_j) \in \mathcal{S}\times \mathcal{S}} E_{S{\times}S}(s_i, s_j) +
  \sum_{f_k\in \mathcal{F}} E_F^k(\mathcal{S}).
\end{eqnarray*}
The first term ($E_S$) is a unary potential, measuring the consistency
between the scan placement and the floorplan image. The second term
($E_{S\times S}$) is a binary potential, measuring the consistency
between pairs of scan placements. The third term ($E_F$)
counts the number of floorplan pixels covered by the scan placements,
and is summed over floorplan image pixels $\mathcal{F}$.
%
$E^k_F$ has a form of a higher order term, but will be approximated by a
sum of pairwise terms as explained below.
The first ($E_S$) and the third ($E_F^k$) terms are the main
contributions, while the second term ($E_{S{\times}S}$)
resembles existing approaches.

\subsection{Scan-to-floorplan consistency potential}

Scan-to-floorplan consistency ($E_{\mathcal{S}}$) needs to overcome
vastly different modalities between a floorplan image and real sensor
data.  Our measure is the sum of the semantic and geometric
penalties (See Fig.~\ref{fig:scan-to-floor}).

\mysubsubsection{Semantic cue} The semantic cues exploit door
detections.
The door detection are used to align pairs of scans in a recent
work by Yan et al.~\cite{yan2016block}. We use door detections to align
a 3D scan against a floorplan image. Door detection results are
represented as a set of pixels in a floorplan image or per-scan evidence
images.
The semantic penalty is defined for every door-pixel in the evidence
image. Let $f_p$ denote the corresponding floorplan pixel under the scan
placement. The penalty is 0 if $f_p$ is also a door-pixel.  Since not
all the doors are marked in a floorplan, $f_p$ may be an unmarked
door-pixel. In that case, $f_p$ must be in a door-way (i.e., white
pixel). Therefore, if $f_p$ is not a door pixel and has intensity more
than 0.4, we set the penalty to 0.5, otherwise 1.0.  The average
penalty over all the door-pixels in the evidence image is the
semantic penalty.





\mysubsubsection{Geometric cue} Measuring the consistency between the
floorplan image and the point evidence image is a real challenge:
1) A floorplan image contains extra symbols that are not in evidence
images; 2) An evidence image contains objects/clutter
that are not in a floorplan image; 3) The style of a floorplan
(e.g., line thickness) may vary; and 4) Both are essentially
line-drawings, making the comparison sensitive to small errors.
%
In practice, we have found that the following consistency potential
provides a robust metric.

We first apply a standard morphological dilation operation (as a
gray-scale image) to a floorplan image, using the OpenCV default
implementation with a $5 \times 5$ kernel. We then measure how much of
the point evidence image is NOT explained by the floorplan, by 1)
subtracting the dilated floorplan image $(DF)$ from the point evidence
image $(P)$; then 2) clamping the negative intensities to 0. The sum of
intensities in this residual image $(max(0, P - DF))$ divided by the sum
of intensities in the original evidence image $(P)$ calculates the
amount of the discrepancy. We swap the role of the floorplan and a point
evidence image, compute the other discrepancy measure, and take the
average.








%

\subsection{Scan-to-scan consistency potential}

Different from standard MRF formulation, we do not know which pairs of
variables (i.e., scans) should have interactions, because our variables
encode the placements of the scans. Therefore, we set up a potential for
every pair of scans. The potential measures the photometric and
geometric consistencies between the two scans given their
placements. The photometric consistency uses Normalized Cross
Correlation of local image patches. The
geometric consistency measures the discrepancy between the point and
free-space evidence information stored over the voxel grids. The
consistency measures are based on standard techniques, and we refer the
details to the supplementary material.

\subsection{Floorplan coverage potential}


The third potential seeks to increase the number of floorplan pixels
covered by the scan. This acts as a counter-force against the
scan-to-scan consistency potential, which has a strong bias to stack
multiple scans at the same location, because 1) this potential was added
for every pair of scans; and 2) the potential goes down only when scans
overlap.

The floorplan coverage potential can be implemented by the sum of the
sub-potentials over the floorplan pixels, each of which returns $1$
if the pixel is not covered by any scans, otherwise $0$.
We define that ``a scan covers a floorplan pixel'' if the pixel is
inside the free-space mask. This sub-potential depends on any scan, one
of whose placement candidates covers the pixel, and usually becomes
higher-order. In practice, most rooms are scanned only once or twice,
and the approximation by pairwise terms has worked well.
More precisely, for every floorplan pixel, we identify a set of scans,
one of whose placement candidates covers the pixel. For every pair
of such scans, we form a pairwise potential that becomes 0.0 if exactly
one of the scans covers the pixel (ideal case), $0.5$ if both cover the
pixel, and $1.0$ if none covers.


%% file: algorithm.tex
\section{Inference} \label{section:optimization}

Naive inference will be infeasible to solve our MRF problem. The label
space is massive (i.e., 4 rotations x 50 million translations per
variable). The energy is not sub-modular. The key insight is that while
indoor scenes are full of repetitions, there are not too many places or
rooms that have exactly the same surrounding geometry and door
placements.
Therefore, simply identifying significant (negative) peaks in the unary
potential can restrict a set of feasible placements for each scan.
%
%
%
In practice, exhaustive evaluation of all the unary costs are still
infeasible (200 million possible placements per scan),
%
%
%
and we employ a standard hierarchical search scheme to identify a small
number (5 in our experiments) of placement candidates per scan.

The hierarchical search scheme works as follows. First, we build an
image pyramid of 5 levels for each floorplan image, an evidence image,
or a door-detection image. Second, we exhaustively evaluate all the
unary costs at the top level, and keep all the local minima less than a
threshold. Then, by level by level, we iterate evaluating the unary
costs at the children pixels under the current local minima, and
applying non-local min suppression with a thresholding.
%
This hierarchical search runs for each of the four
orientations. The best five placements at the bottom level of these
searches are reported as the candidates.

While this search strategy is relatively straightforward, a few
algorithmic details are worth noting. First, we use the maximum (minimum
for a floorplan) intensity over $2\times 2$ pixels instead of the
averaging in the image pyramid creation, as images are near
binary. Second, the non-local min suppression looks at a much larger
area than the direct neighbors, as the function tends to be peaky. Let
$W_B$ and $H_B$ be the width and height of the tight bounding box
containing the floorplan mask. We look at a square region whose size is
$(W_B + H_B)/80$.
Third, the threshold at the non-local min suppression is the mean minus
the standard deviation of the evaluated scores at the same pyramid
level.
Fourth, we speeded up the unary potential evaluation by skipping scan
placements when more than 30\% of the corresponding free-space mask goes
outside the building-mask.
Lastly, $7\times 7$ children pixels (every other pixel is chosen
at the perimeter for speed) instead of $2\times 2$ are searched under
each local minimum for more robustness.

With five placement candidates per scan, we resort to the
tree-reweighted message passing
algorithm~\cite{kolmogorov2006convergent} to optimize our non-submodular
energy. Each variable is initialized as the placement with the best
unary potential. The optimization usually converges after $50$
iterations.

%% file: results.tex
\section{Experimental results and discussions} \label{section:results}

\begin{figure*}[htb]
  \includegraphics[width=\textwidth]{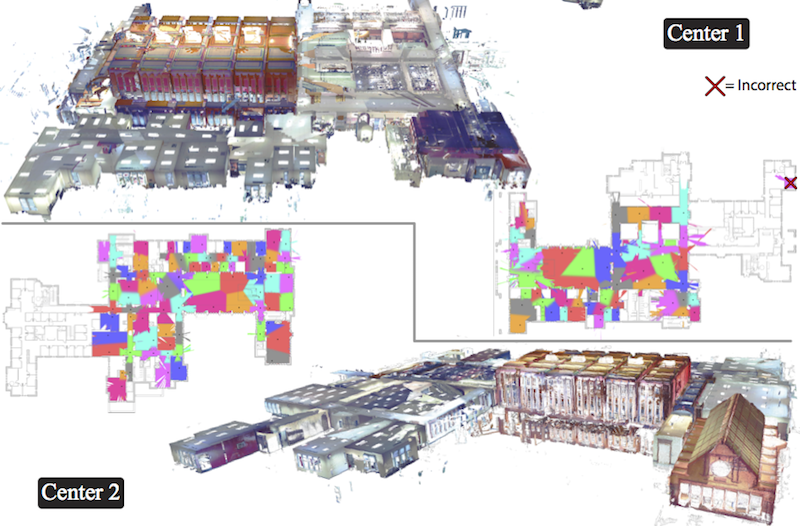}
 \vspace{-0.62cm}
 \caption{Placement results for Center1 and Center2. A merged
3D point-cloud, and 2D colored free-space masks are shown.
 }\label{fig:center}
\end{figure*}

\begin{figure*}[htb]
  \includegraphics[width=\textwidth]{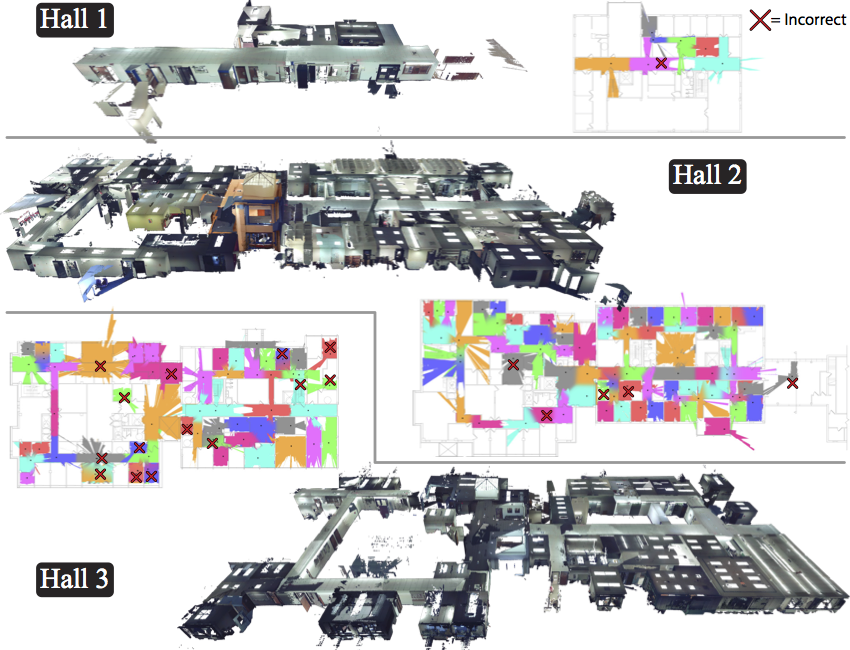}
 \vspace{-0.62cm}
 \caption{Placement results for Hall1, Hall2, and Hall3.} \label{fig:hall}
\end{figure*}

\newcommand{\worst}[1]{\textcolor{red}{#1}}
\newcommand{\best}[1]{\textcolor{cyan}{#1}}
\newcommand{\second}[1]{{#1}}


\begin{figure*}
  \includegraphics[width=\textwidth]{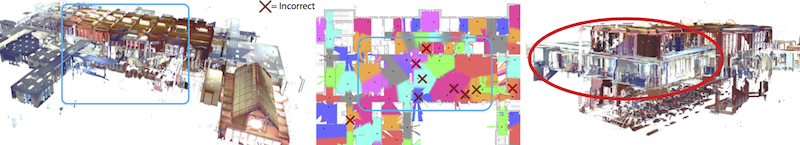}
 \vspace{-0.62cm}
  \caption{Results when replacing our unary potential with a distance
 transform on Center 2.  Major placement errors occur in the middle of a
 big hall. The right cutaway point rendering shows the magnitude of the
 error.}\label{fig:DT}
\end{figure*}

\begin{figure}[tbh]
  \centering
  \includegraphics[width=0.9\columnwidth]{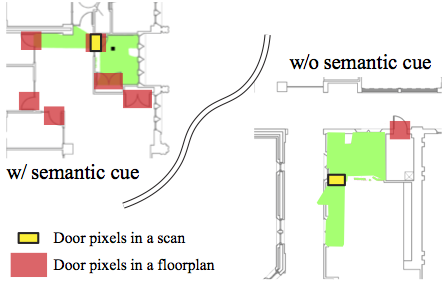}
 \caption{The semantic cue (i.e., door detection) resolves
 ambiguities. The figure shows the best placement based on the unary
 potential with or without the semantic cue.
 }
 \label{fig:door}
\end{figure}

\begin{figure}[tbh]
 \includegraphics[width=\columnwidth]{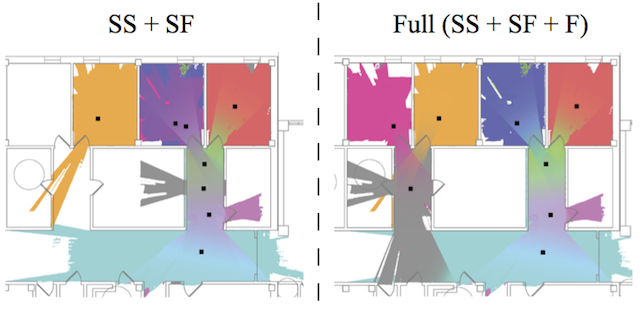}
 \caption{Final scan placements with or without the floorplan coverage
 potential, which mitigates the stacking bias visible on the left.}
 \label{fig:stacking}
\end{figure}

We have used C++ for implementation and Intel Core I7 CPU with 16GB
RAM PC. Three computational expensive steps are preprocessing,
unary-potential evaluation, and TRW optimization, where the running time
is roughly proportional to the number of the input scans. For large
datasets with 70 to 80 scans, these steps roughly take 5 hours, 2.5
hours, and 30 minutes, respectively.
The preprocessing is the bottleneck due to I/O and processing of the
massive scan files, which can be parallelized if necessary.

\begin{table}[tb]
  \centering
  \begin{tabular}{r||r|r|r}
   Data & SF & SF+SS & All (SF+SS+F) \\ \hline
   Center 1 & $\worst{12\%}$ & $\best{0\%}$ & $2\%$  \\
   Center 2 & $3\%$ & $2\%$  & $\best{0\%}$ \\
   Hall 1 & $\worst{29\%}$   & $\best{15\%}$ & $\best{15\%}$ \\
   Hall 2 & $12\%$  & $15\%$  & $\best{7\%}$   \\
   Hall 3 & $44\%$  & $47\%$ & $\best{34\%}$  \\ \hline
  \end{tabular}
 \vspace{-0.2cm}
  \caption{To assess the contributions of each potential, we have run
 our algorithm with different combinations of the potentials. SF,
 SS and F denotes the scan-to-floorplan, scan-to-scan, and floorplan
 coverage potentials, respectively. For example, the left column shows
 the error rate when only the unary (SF) potential is used.
 }\label{table:stat1}
\end{table}

\begin{table*}[tb]
  \centering
  \begin{tabular}{r|r||r|r||r|r|r||r|r|r||r|r|r}
   \multicolumn{2}{r||}{} & \multicolumn{2}{c||}{Naive SSD} &
   \multicolumn{3}{c||}{SSD} & \multicolumn{3}{c||}{Distance transform}
   & \multicolumn{3}{c}{Ours (unary)} \\ \cline{3-13}
  & & & & & & & & &  \\[-6pt]
  Name & \#Scans & Top 1 & Top 5 & Top 1 & Top 5 & Final & Top 1 & Top 5
   & Final & Top 1 & Top 5 & Final \\ \hline

  Center 1 & $50$ & $\worst{96\%}$ & $\worst{94\%}$ & $24\%$ & $12\%$ & $\worst{12\%}$ &
 $\second{14\%}$ & $4\%$ & $\worst{12\%}$ & $\best{6\%}$ & $\best{2\%}$ & $\best{2\%}$ \\

  Center 2 & $80$ & $\worst{94\%}$ & $\worst{91\%}$ & $\second{18\%}$ & $10\%$
   & $10\%$  & $20\%$ & $12\%$ & $\worst{13\%}$ & $\best{1\%}$ & $\best{0\%}$ & $\best{0\%}$ \\

  Hall 1 & $7$ & $\worst{100\%}$& $\worst{100\%}$ & $43\%$ & $\best{0\%}$ & $\best{15\%}$ &
   $\second{29\%}$ & $29\%$ & $\worst{29\%}$ & $\best{15\%}$ & $\best{0\%}$ & $\best{15\%}$ \\

  Hall 2 & $75$ & $\worst{100\%}$ & $\worst{94\%}$ & $48\%$ & $26\%$ & $\worst{37\%}$ &
   $\best{12\%}$ & $\best{8\%}$ & $16\%$ & $\second{22\%}$ & $10\%$ & $\best{7\%}$ \\

  Hall 3 & $65$ & $\worst{97\%}$ & $\worst{96\%}$ & $71\%$ & $56\%$ & $\worst{61\%}$ &
   $\best{45\%}$ & $22\%$ & $37\%$ & $\best{45\%}$ & $\best{32\%}$ & $\best{34\%}$ \\ \hline
  \end{tabular}
  \caption{We have compared our results against several image distance
 metrics that replace the scan-to-floorplan (unary) potential.
 Columns ``Top 1'' and ``Top 5'' indicate if the ground-truth is in the
 top 1 or 5 placements based solely on the unary potential.
 Column ``Final'' reports the error rate after the MRF
 optimization with the replaced unary potential.
 Naive SSD simply takes the sum of squared differences between the
 floorplan image and the point-evidence image. SSD computes the same
 measure but only inside the free-space evidence mask.
 Distance transform makes the floorplan image into binary with a
 threshold 0.4, constructs its distance-transform
 image~\cite{felzenszwalb2004distance}, then takes the sum of element-wise
 product with the point-evidence image inside the free-space evidence
 mask. \yasu{put red, cyan for top 5 and final, too.}
 }\label{table:comp}
\end{table*}

Figures~\ref{fig:center} and \ref{fig:hall} show our main results. For
each dataset, we first show the merged point cloud. The visualization of
the scan placements are given by colored floorplan-masks.  When multiple
masks cover the same pixel, the color of the closest scan is chosen.

We have manually inspected every result to check the placement
correctness, where erroneous ones are highlighted in the
figure.
\yasu{Erik, please put some markers showing which ones are incorrect
inside the colorful freespace mask in fig 5 and fig 6.}
Table~\ref{table:stat1} shows our placement error rates (i.e., the ratio
of incorrectly placed scans). Our algorithm has
successfully aligned most of the scans.
We have not scanned the right wing of the building in Center 1 and
Center 2 (See Fig.~\ref{fig:input}), which makes a large space for scans
to be misplaced. Nonetheless, our method has only one misplacement in
that area (Center 1).
Note that Hall 3 is an exception in which we make many errors due to the
glitch in the floorplan image. We will discuss failure cases later.

Our MRF formulation consists of the three potentials. We have run our
algorithm with a few different combinations to assess their
contributions. Table~\ref{table:stat1} shows relatively low error rates
for the unary-only results (SF) and demonstrates the power of utilizing
floorplans.
%
The table also shows that the standard pairwise potential (SS), the main
cue for existing approaches, has consistently improved the alignment
accuracy for easier datasets (Center 1, Center 2, and Hall1), but not
for the harder two cases. The floorplan coverage potential is crucial
for challenging datasets (Hall2 and Hall3), which are full of
repetitions and ambiguity.

Due to the large label space, our approach has required the unary
potential to limit the feasible label-set.
Nonetheless, we have taken the limited label-space and run our algorithm
without the unary term, whose results are much worse than the unary-only
solution. \yasu{check if this is the case.} \erik{What do you mean
without unary?  Removing unary from the MRF formulation?}
Furthermore, we have experimented the feasibility of conventional
scan-to-scan alignment techniques, in particular, Autodesk ReCap
360~\cite{autodesk_recap} and K-4PCS~\cite{theiler2014fast}, which do not use floorplan
data. Both methods have failed to generate any type of meaningful
result, again confirming the importance of utilizing a floorplan image
for our problems.
Note that for fair comparison, we have evaluated the fully automated
mode in Autodesk ReCap 360. We have also utilized its interactive mode
in aligning the scans, but the process has been extremely painful and
time-consuming (6 or 7 hours of intensive mouse clicking for large
dataset). Furthermore, the final alignments have suffered from major
errors due to the forced automatic refinement, which cannot be avoided.
Table~\ref{table:stat1} also shows that the standard pairwise potential
(SS) has consistently improved the alignment accuracy for easier
datasets (Center 1, Center 2, and Hall1). Lastly, the floorplan coverage
potential is crucial for challenging datasets (Hall2/Hall3), which
are full of symmetry and ambiguity.

To further evaluate the contribution of our unary potential and the
effectiveness of utilizing a floorplan image, we have experimented with
three alternative image matching metrics to replace the unary term (See
Table~\ref{table:comp}). The same hierarchical search scheme
(Sect.~\ref{section:optimization}) has been used.
%
The naive SSD without the mask fails badly as expected. SSD and Distance
transform utilizing our masks has achieved reasonable accuracy, which is
remarkable, considering the fact that the pairwise scan alignment
without floorplan (i.e., current state-of-the-art) has completely failed
in all the examples.
%
Columns ``Top 1'' and ``Top 5'' indicate if the ground-truth placement
is in the top 1 or 5 candidates based solely on the unary potential. It
is worth noting that expanding the candidate list did not help in
reducing the error rate for Top 5, because failure cases are usually
extreme.

Figure~\ref{fig:DT} illustrates typical failure modes of Distance
transform, which tends to concentrate scans in large rooms.  Our
analysis is that a large room tends to have non-architectural lines or
symbols to fill-in an empty canvas, which makes the distance transform
image contain small values, and allow lower-energy placements.
%
%


Figure~\ref{fig:door} illustrates the effects of the semantic cue in the
unary potential (i.e., the door detection). Indoor scenes are full of
symmetries and repetitions, which makes the comparison of pure geometry
(i.e., geometric cue) susceptible to local minima. The figure
demonstrates a representative case, where the door detections break such
an ambiguity.

When the placement is ambiguous even with the geometric and the semantic
cues, we rely on the MRF optimization with the full three
potentials. Figure~\ref{fig:stacking} compares the final scan placements
with or without the floorplan coverage potential.
The floorplan coverage potential seeks to avoid ``stacking'' and evenly
distribute the placements.
%





Our method is not perfect and has exposed several failure modes. First,
our approach tends to make mistakes for small storage-style rooms, where
a small room with a lot of clutter makes the geometric cue very noisy.
Second, there are genuinely ambiguous cases where the scene geometry,
appearance, and door locations are exactly the same. Lastly, our method
has made major errors in Hall 3, simply because the floorplan has not
reflected renovations in the past.
Unfortunately, it was difficult to identify erroneous scans based on the
potentials. At the presence of problematic scans, the MRF optimization
seems to shuffle around scan placements including correct ones to
achieve a low energy state.  Nonetheless, the total potential, in
particular, the magnitude of the total potential divided by the number
of scans is a good indicator of success. The quantity for Hall 3 is a
few times larger than the others and indicates that ``something is
wrong''.
Our main future work is to develop a robust algorithm to detect
potentially erroneous scan placements, which will allow a quick user
feedback to correct mistakes.
We will share our source code and high-end building-scale datasets to
further enhance indoor mapping research.



%% file: conclusion.tex


%% file: acknowledgments.tex
\section{Acknowledgments}

This research is partially supported by National Science Foundation under grant IIS 1540012 and IIS 1618685.